\documentclass[journal]{IEEEtran}
\usepackage{amsmath,amsfonts}
\usepackage{algorithmic}
\usepackage{array}
\usepackage[caption=false,font=normalsize,labelfont=sf,textfont=sf]{subfig}
\usepackage{textcomp}
\usepackage{stfloats}
\usepackage{url}
\usepackage{verbatim}
\usepackage{graphicx}

\def\BibTeX{{\rm B\kern-.05em{\sc i\kern-.025em b}\kern-.08em
    T\kern-.1667em\lower.7ex\hbox{E}\kern-.125emX}}
\usepackage{balance}

\usepackage{epstopdf}

\usepackage{soul}
\soulregister\ref7

\usepackage[table,xcdraw]{xcolor}

\usepackage{multirow}

\usepackage{tabularx}
\usepackage{array}
\usepackage{cite}
\usepackage{hyperref} 
\hypersetup{
colorlinks=true,
linkcolor=red,
citecolor=green,
urlcolor = red
}

\usepackage{makecell}

\begin{document}
\title{Underwater Image Enhancement with Cascaded Contrastive Learning}
\author{Yi Liu, Qiuping Jiang,~\IEEEmembership{Senior Member,~IEEE}, Xinyi Wang, Ting Luo,~\IEEEmembership{Member,~IEEE},\\Jingchun Zhou,~\IEEEmembership{Senior Member,~IEEE}
\thanks{This work was supported in part by the Natural Science Foundation of Zhejiang under Grant No. LR22F020002, in part by the Natural Science Foundation of China under Grant No. 62271277, and in part by the Natural Science Foundation of Ningbo under Grant No. 2022J081. (\textit{Corresponding author: Qiuping Jiang})}
\thanks{Y. Liu and T. Luo are with the College of Science and Technology, Ningbo University, Ningbo 315300, China. (e-mail: lewis081@126.com; luoting@nbu.edu.cn).}
\thanks{Q. Jiang is with the School of Information Science and Engineering, Ningbo University, Ningbo 315211, China. (e-mail: jiangqiuping@nbu.edu.cn).}
\thanks{X. Wang is with the Hangzhou DtDream Science and Technology Ltd., Hangzhou 310024, China. (e-mail: 13738100147@163.com).}
\thanks{J. Zhou is with the College of Information Science and Technology, Dalian Maritime University, Dalian 116026, China. (e-mail: zhoujingchun@dlmu.edu.cn).}
}

\markboth{IEEE Transactions on Multimedia}
{How to Use the IEEEtran \LaTeX \ Templates}

\maketitle

\vspace{-2em}
\begin{abstract}
\small Underwater image enhancement (UIE) is a highly challenging task due to the complexity of underwater environment and the diversity of underwater image degradation. Due to the application of deep learning, current UIE methods have made significant progress. Most of the existing deep learning-based UIE methods follow a single-stage network which cannot effectively address the diverse degradations simultaneously. In this paper, we propose to address this issue by designing a two-stage deep learning framework and taking advantage of cascaded contrastive learning to guide the network training of each stage. The proposed method is called CCL-Net in short. Specifically, the proposed CCL-Net involves two cascaded stages, i.e., a color correction stage tailored to the color deviation issue and a haze removal stage tailored to improve the visibility and contrast of underwater images. To guarantee the underwater image can be progressively enhanced, we also apply contrastive loss as an additional constraint to guide the training of each stage. In the first stage, the raw underwater images are used as negative samples for building the first contrastive loss, ensuring the enhanced results of the first color correction stage are better than the original inputs. While in the second stage, the enhanced results rather than the raw underwater images of the first color correction stage are used as the negative samples for building the second contrastive loss, thus ensuring the final enhanced results of the second haze removal stage are better than the intermediate color corrected results. Extensive experiments on multiple benchmark datasets demonstrate that our CCL-Net can achieve superior performance compared to many state-of-the-art methods. In addition, a series of ablation studies also verify the effectiveness of each key component involved in the proposed CCL-Net. The source code of CCL-Net will be released at \url{https://github.com/lewis081/CCL-Net}.
\end{abstract}


\section{Introduction}
\IEEEPARstart{T}{he} ocean covers 80\% of the earth's surface and holds abundant resources. Exploring marine resources is of great significance for humanity. Underwater tasks such as archaeology \cite{ludvigsen2007applications}, marine biology research \cite{shi2022detecting, Cheng2023BidirectionalCM}, and equipment inspections \cite{Sun2020InverseSA, Ahn2020AnOI} largely rely on high-quality visual information conveyed by underwater images \cite{zhou2024decoupled, jiang2022single, chen2024perception}. However, underwater images usually suffer from diverse degradation issues such as color cast, hazy effect, blurred details, and low light, etc, due to the light absorption and scattering by the water media\cite{jaffe2014underwater, Zhou2023UnderwaterCI, Jiang2022UnderwaterIE, Zhuang2022UnderwaterIE, chen2019reference, yi2024no, jiang2022unsupervised, zhou2024dtkd}. 
Therefore, it is highly essential to devise effective underwater image enhancement (UIE) methods to improve the visual quality of real-world raw underwater images, enabling better exploration and understanding of the marine world for humanity.

Recently, a considerable number of deep learning-based UIE methods \cite{TiesongUIE,Jiang2023PerceptionDrivenDU,li2020underwater,li2019underwater,li2021underwater, Zhou2024HCLRNetHC} have been proposed to enhance underwater image quality. However, the issues of underwater image degradation have not been fully resolved due to the diversity and complexity of underwater environment. Since the characteristics of different underwater degradation types are not the same, the single-stage networks usually resort to constrain the entire network by multiple losses defined in different perspectives. Nevertheless, it is hard to achieve the optimal balance for these losses, making the single-stage networks difficult to effectively address multiple degradation issues simultaneously.
Additionally, some UIE methods have achieved promising results by using multi-branch strategies\cite{zhang2022dual,jiang2023five,Hu2021TwoBranchDN} to address color cast and hazy effect, separately. However, they did not impose explicit constraints in dealing with these two primary degradation issues, failing to offer specific guidance for subsequent individual optimization.

\begin{figure}[!t]
\centerline{\includegraphics[width=0.9\linewidth]{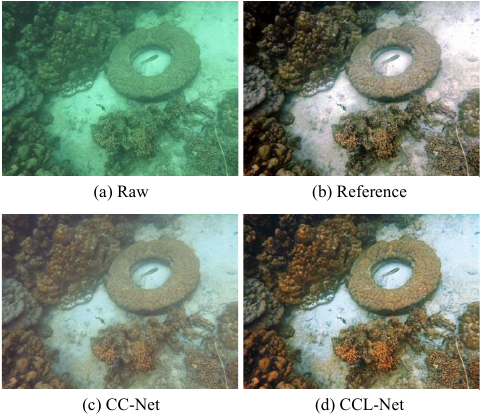}}
\caption{The enhanced results by proposed CCL-Net. (a) is the raw underwater images with a predominant greenish tone. (b) is the pseudo-reference counterpart. (c) is the output of the CC-Net. (d) is the output of the HR-Net, i.e., the final result of CCL-Net.}
\label{fig_intro}
\end{figure}

Moreover, most existing UIE methods \cite{li2020underwater,li2019underwater,li2021underwater,qi2022sguie,wang2022semantic,wu2021two} usually employ well-enhanced images as pseudo-reference images (i.e., positive samples) to train the enhancement network. However, these approaches can only encourage the network output closer to the upper bound while without ensuring the lower bound, meaning that it cannot guarantee the worst case. Inspired by the recent application of contrastive learning \cite{wu2021contrastive}, raw images and reference images are respectively utilized as the negative samples and positive samples to establish a contrastive loss for constraining the network training process. This approach encourages restored images close to positive samples and away from negative samples. For UIE, it is also crucial for the network to maintain a reasonable lower bound so that the enhancement result does not deteriorate excessively. Further, different quality of positive and negative samples can lead to different upper and lower bound. Therefore, how to adequately and reasonably utilize positive and negative samples to constrain and guide the enhancement network training is highly desired.

To address the above two issues, we propose a tailored UIE approach based on cascaded contrastive learning (CCL). The proposed method called CCL-Net in short contains two cascaded sub-networks (i.e., a color correction sub-network called CC-Net and an haze removal sub-network called HR-Net) with each sub-network constrained by a contrastive loss and other conventional losses simultaneously. Inspired by the divide-and-conquer strategy, the two cascaded sub-networks, i.e., CC-Net and HR-Net, are respectively designed to address the color cast and hazy effect issues. 
We consider that the network can achieve a better lower bound if the negative samples with better quality are used to build the contrastive loss. This is the key factor to build contrastive loss. Therefore, we construct two contrastive losses in our two-stage framework, and the negative samples in these two contrastive losses are different.
For the CC-Net, in addition to a tailored color loss defined in the Lab color space, we also construct a contrastive loss by taking the input raw underwater image as negative samples and the pseudo-reference images as positive samples in the RGB color space to further improve the color correction capability. As shown in Fig. \ref{fig_intro} (c), the color cast issue is largely resolved, leaving mostly the hazy effect. Based on the cascaded way of our two-stage framework, we can better exert the advantage of the contrastive learning in the second stage and make the HR-Net achieve a better lower bound. Therefore, in the HR-Net, in addition to an conventional loss, we also introduce a contrastive loss by taking the enhanced result from the CC-Net as negative samples and the pseudo-reference images as positive samples to further improve overall performance. As shown in Fig. \ref{fig_intro} (d), the hazy effect is well removed compared with the one shown in Fig. \ref{fig_intro} (c). In a nutshell, the main contributions of this paper are as follows:
\begin{itemize}
    \item We propose a two-stage cascaded network architecture with each sub-network specifically designed to address the color cast and hazy effect issues of underwater images, respectively. In view of the function of each stage, we impose tailored losses on the corresponding sub-network to effectively facilitate independent sub-network optimization for distortion-specific enhancement. 
    \item We adopt a new training paradigm that combines the two sub-networks and cascaded contrastive loss to incrementally improve the lower and upper bound of the enhancement result. By introducing contrastive loss and carefully utilizing negative samples, the network achieves a better lower bound of the output. This approach fully explores the advantages of the two-stage framework and contrastive loss.
    \item Extensive experiments on multiple benchmark datasets demonstrate that our proposed CCL-Net achieves superior performance in terms of both visual quality and quantitative metrics. Besides, a series of ablation studies also verify the effectiveness of each key component involved in the proposed CCL-Net.
\end{itemize}

\section{Related Work}
\subsection{Deep Learning-based Methods}
Deep learning initially showcased its impressive capabilities and garnered attention in the field of image classification. In 2012, AlexNet\cite{krizhevsky2012imagenet} achieved the top position in the ImageNet\cite{deng2009imagenet} competition for that year. Following that, VGG\cite{simonyan2014very} in 2014 and ResNet\cite{he2016deep} in 2015 continued to lead advancements in image classification. During this time, an increasing number of tasks began to explore the use of deep learning to address their specific challenges. In recent years, deep learning has also been applied to address the issue of UIE.

In early attempts, Mao et al. \cite{mao2016image} introduced deep learning methods into the field of UIE byleveraging the achievement of deep learning in image classification and its powerful nonlinear fitting capability. They employed a deep convolutional autoencoder network with skip connections from residual networks for UIE, achieving optimal results across multiple databases. Wang et al. \cite{wang2017deep} introduced the UIE-Net based on a physical model of underwater imaging. Li et al. \cite{li2020underwater} trained 10 UWCNN models tailored to various underwater scenes for UIE. The lightweight network also allows real time enhancement of underwater videos. Also, Li et al. \cite{li2019underwater} construct the first real-world underwater paired image dataset for end-to-end supervised training and devised the Water-Net for deep UIE. This approach yielded promising results when applied to enhance the quality of real-world underwater images. More recently, Li et al. \cite{li2021underwater} introduced the UColor network with multi-color space embedding. This method incorporates the inverse transmission map as the attention map to guide the decoding network to pay more attention to those regions with more severe degradation and thus better enhance these areas accordingly. Fu et al. \cite{fu2022uncertainty} proposed a novel probabilistic network called PUIE for UIE by combining variational autoencoder and consensus process. It copes with reference map ambiguity and bias, yielding robust enhancement predictions comparable to state-of-the-art methods. Guo et al. \cite{guo2023underwater} proposed a simple U-shape UIE network called NU2Net which can obtain promising performance when coupled with the pre-trained UIQA URanker as additional supervision.

Although the impressive progress has been achieved by the existing methods, it should be noted that merely employing single-stage network without intermediate explicit constraints fails to effectively address both the color cast and haze effect issues. By contrast, we in this work propose a cascaded two-stage framework that addresses these two degradation issues in a divide-and-conquer manner. Specifically, the first stage addresses the color distortion issue via explicit color constrained loss while the second stage focuses on removing the hazy effect.

\subsection{Contrastive Learning}
Contrastive learning has achieved great success in self-supervised representation learning\cite{henaff2020data, tian2020contrastive,sermanet2018time,chen2020simple,he2020momentum} by constructing a loss function that guides the network to bring the positive samples getting closer and the negative samples being farther away in the latent space.

In the early stage, self-supervised contrastive learning exhibited robust competitiveness in high-level visual tasks. Chen et al. \cite{chen2020simple} introduced the SimCLR framework, which used contrastive learning to obtain visual feature representation, achieving superior accuracy on ImageNet compared to existing self-supervised and semi-supervised methods. He et al. \cite{he2020momentum} proposed the Momentum Contrast (MoCo) method for unsupervised visual representation learning. MoCo leverages a dynamic queue and moving average encoders to facilitate non-supervised contrastive learning, outperforming supervised methods across seven detection and segmentation tasks.

For the low-level vision tasks, Wu et al. \cite{wu2021contrastive} introduced AECR-Net to address image dehazing. To enhance the effectiveness of network training with hazy images, a contrastive loss was established using clear and hazy images as positive and negative samples, respectively. This guided the network to generate outputs closer to clear images and farther from hazy images, ensuring a controlled lower bound of the network's outputs and preventing uncontrolled worst-case situation. Wang et al. \cite{wang2022ucl} proposed UCL-Dehaze which explored contrastive learning with an adversarial training effort to leverage unpaired real-world hazy and clean images to reduce the domain shift between synthetic and real-world haze. For UIE, Liu et al. \cite{liu2022twin} introduced an object-guided twin adversarial contrastive learning method which enhanced the visual quality of raw underwater images while improving the accuracy of object detection.

Despite the notable advancement in contrastive learning for high-level visual tasks, its exploration in low-level visual tasks remains in its infancy. For UIE, we believe that employing contrastive loss will also be beneficial. The naive way to use contrastive learning for UIE is to directly take the raw images as negative samples and the pseudo-reference images as positive samples to build the contrastive loss. Different from the naive way, we in this work propose a two-stage framework with cascaded contrastive learning for UIE, which uses the enhanced result of the first stage as negative samples for building the contrastive loss in the second stage to encourage the raw underwater image to be progressively enhanced.

\begin{figure*}[!t]
\centerline{\includegraphics[width=\linewidth]{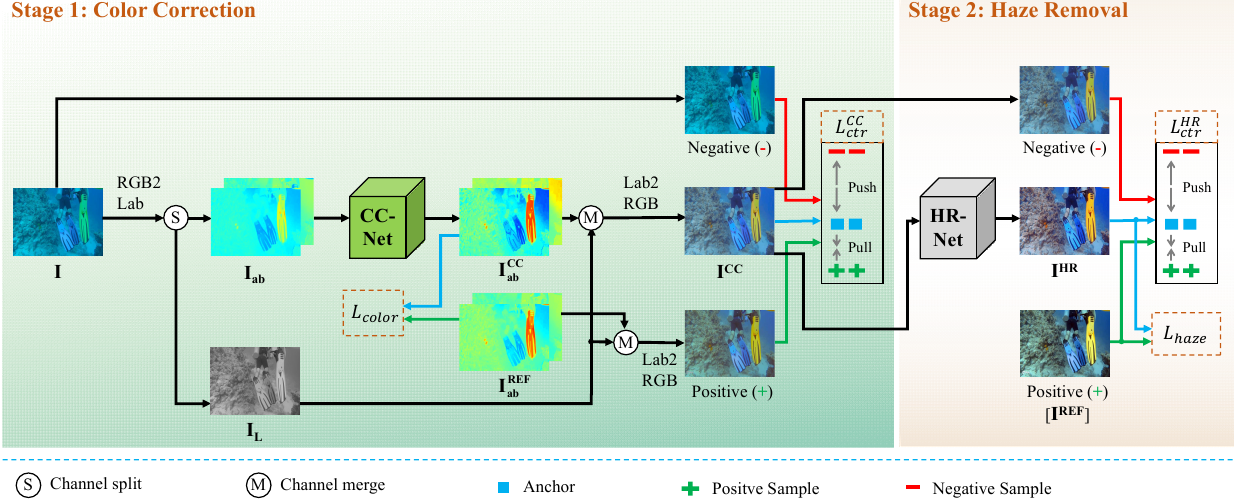}}
\caption{Architecture of the proposed CCL-Net. It consists a color correction stage and a haze removal stage. In the first color correction stage, a raw underwater image $\mathbf{I}$ is first transformed into the Lab color space, resulting in a luminance channel $\mathbf{I_L}$ and two color channels $\mathbf{I_{ab}}$. Then, the luminance channel $\mathbf{I_L}$ is discarded before feeding into the CC-Net which focuses on color correction only. As a result, only the two color channels $\mathbf{I_{ab}}$ are fed into the CC-Net, resulting in corrected color channels $\mathbf{I_{ab}^{CC}}$. Finally, the corrected color channels $\mathbf{I_{ab}^{CC}}$ and the original luminance channel $\mathbf{I_L}$ are merged and converted back into a color corrected yet hazy underwater image in the RGB format $\mathbf{I^{CC}}$. A hybrid loss consisting of an explicit color correction loss and a contrastive loss is used as the constraint of CC-Net. In the second haze removal stage, the color corrected yet hazy underwater image from the previous stage $\mathbf{I^{CC}}$ is fed into the HR-Net, resulting in an enhanced result with clear structures and fine details. Similarly, a hybrid loss consisting of an explicit structural similarity (SSIM) loss and a contrastive loss is used as the constraint of HR-Net. This helps the enhanced image approach the reference image and get away from the color corrected yet hazy image, ultimately achieving a visually appealing enhanced image with high color saturation and contrast.}
\label{fig_overview}
\end{figure*}

\section{Proposed Method}
\subsection{Overview}
The architecture of our proposed CCL-Net is shown in Fig. \ref{fig_overview}. It consists of two stages including a color correction stage and a haze removal stage. The main idea of our method is to address the color cast issue and reduced visibility issue with the CC-Net and HR-Net, respectively. 
Since each sub-network only addresses one degradation issue, the structure of these two sub-networks can be simple.
% In the first color correction stage, a raw underwater image $\mathbf{I}$ is first transformed into the Lab color space, resulting in a luminance channel $\mathbf{I_L}$ and two color channels $\mathbf{I_{ab}}$. Then, the luminance channel $\mathbf{I_L}$ is discarded before feeding into the CC-Net which focuses on color correction only. As a result, only the two color channels $\mathbf{I_{ab}}$ are fed into the CC-Net, resulting in corrected color channels $\mathbf{I_{ab}^{CC}}$. Finally, the corrected color channels $\mathbf{I_{ab}^{CC}}$ and the original luminance channel $\mathbf{I_L}$ are merged and converted back into a color corrected yet hazy underwater image in the RGB format $\mathbf{I^{CC}}$. A hybrid loss consisting of an explicit color correction loss and a contrastive loss is used as the constraint of CC-Net. In the second haze removal stage, the color corrected yet hazy underwater image from the previous stage $\mathbf{I^{CC}}$ is fed into the HR-Net, resulting in an enhanced result with clear structures and fine details. Similarly, a hybrid loss consisting of an explicit structural similarity (SSIM) loss and a contrastive loss is used as the constraint of HR-Net. This helps the enhanced image approach the reference image and get away from the color corrected yet hazy image, ultimately achieving a visually appealing enhanced image with high color saturation and contrast.

\subsection{Color Correction Stage}
As shown in the left side of Fig. \ref{fig_overview}, the color correction stage takes a raw underwater image $\mathbf{I}$ as input and produce a color corrected yet hazy underwater image $\mathbf{I^{CC}}$ as output. Specifically, we develop a dedicated color correction sub-network to tackle the color cast issue caused by underwater absorption. 
Mathematically, the overall procedure of the color correction stage can be formulated as follows:
\begin{equation}
\label{eq_split}
\{\mathbf{I_{ab}}  ,\mathbf{I_{L}}\}  = \mathbb{F}_{s} (\mathbb{F}_{RGB\to Lab}(\mathbf{I} ))
\end{equation}
\begin{equation}
\label{eq_infer_cc}
\mathbf{I_{ab}^{CC}}=\mathcal{M}^{CC} (\mathbf{I_{ab}};\theta^{CC})
\end{equation}
\begin{equation}
\label{eq_merge}
\mathbf{I^{CC}}  =  \mathbb{F}_{Lab\to RGB}(\mathbb{F}_{m}(\mathbf{I_{ab}^{CC}}, \mathbf{I_{L}}))
\end{equation}
where $\mathbb{F}_{RGB\to Lab}$ and $\mathbb{F}_{Lab\to RGB}$ denote the color space conversion, $\mathbb{F}_{s}$ and $\mathbb{F}_{m}$ represent the channel split and merge function, respectively, $\mathbf{I_{ab}}$ and $\mathbf{I_{L}}$ stand for the color channels and luminance channel in the Lab color space, $\mathbf{I_{ab}^{CC}}$ is the corrected color channels which is obtained by the CC-Net $\mathcal{M}^{CC}$ parameterized by $\theta^{CC}$. The proposed CC-Net is trained with the constraint of a hybrid loss ${L}^{CC}$ defined as follows:
\begin{equation}
\label{eq_loss_cc}
L^{CC}=L_{color}+\lambda^{CC}{L}_{ctr}^{CC}
\end{equation}
where $L_{color}$ represents an explicit color correction loss that is designed specially for underwater color cast correction, $L_{ctr}^{CC}$ represents a contrastive loss designed for improving both the lower and upper bound of color correction and $\lambda^{CC}$ is empirically set to $0.5$ to balance the importance of these two terms.

\subsubsection{CC-Net}
We consider the color cast issue as a matter of learning the difference $\Delta (\mathbf{I_{ab}})$ between the color channels $\mathbf{I_{ab}}$ of the raw image and the color channels $\mathbf{I_{ab}^{REF}}$ of the reference image. Mathematically, it can be expressed as follows:
\begin{equation}
\label{eq_abgt}
\mathbf{I_{ab}^{REF}} =\mathbf{I_{ab}}\oplus\Delta (\mathbf{I_{ab}})
\end{equation}
where $\oplus$ is the element-wise addition operation. 

\begin{figure}[!t]
\centering
\centerline{\includegraphics[width=3.47in]{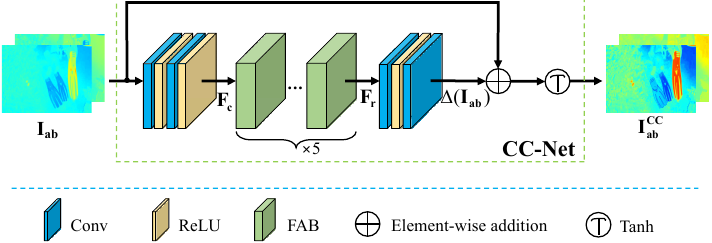}}
\caption{The schematic illustration of CC-Net. The CC-Net learns the difference between raw color channels and reference color channels to achieve color correction.}
\label{fig_cc}
\end{figure}

As illustrated in the Fig. \ref{fig_cc}, the color channels $\mathbf{I_{ab}}$ of the raw image are processed by two serial convolution and ReLU operations, resulting in $\mathbf{F_c}$ as follows:
\begin{equation}
\label{eq_ccnet_f1}
\mathbf{F_{c}}   ={\large (} \mathbb{F}_{ReLU}( \mathbb{F}_{conv}(\mathbf{I_{ab}} )){\large )} _{\times 2}
\end{equation}
where $\mathbb{F}_{conv}$ and $\mathbb{F}_{ReLU}$ represent the convolution and ReLU operation, respectively, $()_{\times 2}$ represents the number of consecutive convolution and ReLU operations. 
Subsequently, $\mathbf{F_c}$ is fed into a series of consecutive Feature Attention Block (FAB)\cite{qin2020ffa} with both spatial and channel attention mechanisms. These blocks iteratively and adaptively select more relevant color representative information $\mathbf{F_{r}}$ in both spatial and channel dimensions as follows:
\begin{equation}
\label{eq_ccnet_f2}
\mathbf{F_{r}}   ={\large (}\mathbb{F}_{FAB}(\mathbf{F_c} ){\large )}_{\times 5} 
\end{equation}
where $\mathbb{F}_{FAB}$ denotes the FAB module \cite{qin2020ffa} and $()_{\times 5}$ represents the number of FAB modules. 
Afterwards, $\mathbf{F_{r}}$ is forwarded into a convolution-ReLU-convolution block to reduce the channel dimensionality to match the color channels. This preserves the most relevant color difference information to generate the $\Delta (\mathbf{I_{ab}})$ as follows:
\begin{equation}
\label{eq_ccnet_deltaab}
\Delta (\mathbf{I_{ab}})=\mathbb{F}_{conv}(\mathbb{F}_{ReLU}(\mathbb{F}_{conv}(\mathbf{F_{r}}))).
\end{equation}

Finally, $\Delta (\mathbf{I_{ab}})$ is added to $\mathbf{I_{ab}}$, followed by a tanh operation $\mathbb{F}_{tanh}$ to restrict the values within the range of [-1, 1]. The output of the CC-Net is the color corrected channels $\mathbf{I_{ab}^{CC}}$ expressed as follows:
\begin{equation}
\label{eq_ccnet_abcc}
\mathbf{I_{ab}^{CC}}=\mathbb{F}_{tanh}(\mathbf{I_{ab}}\oplus\Delta (\mathbf{I_{ab}})).
\end{equation}

From Fig. \ref{fig_intro}(c), it is evident that the color cast issue has been effectively resolved by the CC-Net, with the image primarily exhibiting hazy effect only.

\subsubsection{Hybrid Loss}
As expressed in Eq. \ref{eq_loss_cc}, we will describe the details of $L_{color}$ and $L_{ctr}^{CC}$.

\textbf{Color correction loss}: Existing approaches \cite{zhang2022dual,jiang2023five} have achieved promising results on color enhancement, yet lacking an explicit color constraint, making it unclear whether the network really performs well on color enhancement. Therefore, in this work, we use an explicit color correction loss defined between the enhanced color channels $\mathbf{I_{ab}^{CC}}$ and the referenced color channels $\mathbf{I_{ab}^{REF}}$ as follows:
\begin{equation}
\label{eq_ll2_cc}
L_{color}=\frac{1}{H*W} \left \| \mathbf{I_{ab}^{CC}}-\mathbf{I_{ab}^{REF}}\right \|_{2}^{2}
\end{equation}
where $H$ and $W$ denote the height and width of the image, respectively.

\textbf{Contrastive loss}: In addition to the explicit color loss as the main constraint to guide network training, we introduce an additional contrastive loss to enhance both the lower and upper bound of color correction. The contrastive loss is defined as follows:
\begin{equation}
\label{eq_lc_cc}
{L}_{ctr}^{C C}=\frac{1}{s} \sum_{i=1}^{N}w_{i}\,\frac{\left \| \mathbb{E}_{i}({ \mathbf{I^{CC}})-\mathbb{E}_{i}(\mathbf{\hat{I}^{REF}})}\right \|_{1} }{\left \| \mathbb{E}_{i}(\mathbf{I^{CC}})-\mathbb{E}_{i}(\mathbf{I})\right \|_{1}}
\end{equation}
where $\mathbf{\hat{I}^{REF}}$ is a referenced RGB image which is obtained by merging the raw luminance channel $\mathbf{I_L}$ with the referenced color channels $\mathbf{I_{ab}^{REF}}$ and transforming the merged image to RGB color space, $\mathbb{E}$ is the pre-trained VGG-19 \cite{simonyan2014very} network for feature extraction, $i\in\{1,2,...,n\}$ is the $i$-th layer feature, we select 1st, 3rd, 5th, 9th and 13th layer features of VGG-19, $w_i$ is the corresponding weight and we set $w_1$$\sim$$w_5$ to $1/32$, $1/16$, $1/8$, $1/4$, and $1$, respectively, $s$ is a scaling factor used for normalizing the contrastive loss to the same scale as $L_{color}$ and is empirically set to $100$ in this work.

\subsection{Haze Removal Stage}
As shown in the right side of Fig. \ref{fig_overview}, the haze removal stage takes a color corrected yet hazy underwater image $\mathbf{I^{CC}}$ as input and produces a final enhanced underwater image $\mathbf{I^{HR}}$ as output. Specifically, we develop a dedicated haze removal sub-network to address the hazy effect caused by underwater scattering. Mathematically, the whole procedure of the haze removal stage can be formulated as follows:
\begin{equation}
\label{eq_infer_HR}
\mathbf{I^{HR}}=\mathcal{M}^{HR} (\mathbf{I^{CC}};\theta^{HR}  )
\end{equation}
where $\theta^{HR}$ stands for the parameters of HR-Net model $\mathcal{M}^{HR}$. The proposed HR-Net is trained with the constraint of a hybrid loss ${L}^{HR}$ designed as follows:
\begin{equation}
\label{eq_loss_hr}
{L}^{HR}=L_{haze}+\lambda^{HR}{L}_{ctr}^{HR}
\end{equation}
where $L_{haze}$ represents the structural and textural loss, ${L}_{ctr}^{HR}$ represents the contrastive loss which is specially designed for guaranteeing a better lower bound of haze removal, and $\lambda^{HR}$ is empirically set to $0.5$ to balance the importance of these two terms.

\subsubsection{HR-Net}
A challenging issue in haze removal is the non-uniform distribution of haze concentration. Therefore, we adopt an attention-based multi-scale information fusion module in HR-Net to select appropriate scale features for regions with varying haze concentration.

\begin{figure}[!t]
\centerline{\includegraphics[width=3.47in]{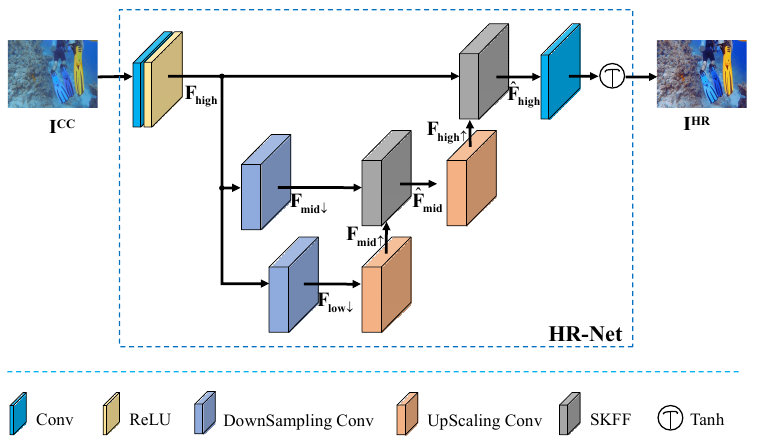}}
\caption{The schematic illustration of HR-Net. An attention-based multi-scale information fusion module is utilized in HR-Net to select appropriate scale features for regions with varying
haze concentration.}
\label{fig_hr}
\end{figure}

As illustrated in Fig. \ref{fig_hr}, the color corrected yet hazy underwater image $\mathbf{I^{CC}}$ first undergoes a convolution layer and a ReLU layer to extract high-resolution feature as follows:
\begin{equation}
\label{eq_hrnet_cr}
\mathbf{F_{high}}=\mathbb{F}_{ReLU}(\mathbb{F}_{conv}(\mathbf{I^{CC}})).
\end{equation}
Then, $2\times$ downsampling convolution layer $\mathbb{F}_{d2}$ and $4\times$ downsampling convolution layer $\mathbb{F}_{d4}$ are applied to $\mathbf{F_{high}}$, yielding medium-resolution feature $\mathbf{F_{{mid}\downarrow}}$ and low-resolution feature $\mathbf{F_{{low}\downarrow}}$ as follows:
\begin{equation}
\label{eq_hrnet_d2}
\mathbf{F_{mid\downarrow}}=\mathbb{F}_{d2} (\mathbf{F_{high}})
\end{equation}
\begin{equation}
\label{eq_hrnet_d4}
\mathbf{F_{low\downarrow}}=\mathbb{F}_{d4} (\mathbf{F_{high}}).
\end{equation}
With the obtained multi-scale features, fusion of different scale features starts from the bottom layer and progresses upwards to the top layer. The bottom-layer feature $\mathbf{F_{low\downarrow}}$ is passed through $2\times$ upscaling convolution layer $\mathbb{F}_{u2}$ to obtain the medium-resolution feature $\mathbf{F_{mid\uparrow}}$ as follows:
\begin{equation}
\label{eq_hrnet_u2_1}
\mathbf{F_{mid\uparrow}}=\mathbb{F}_{u2} (\mathbf{F_{low\downarrow}}).
\end{equation}
In order to fuse adjacent scale features and select most representative features, the medium-resolution fused feature $\mathbf{\hat{F}_{mid}}$ can be obtained as follows:
\begin{equation}
\label{eq_hrnet_skff_1}
\mathbf{\hat{F}_{mid}}=\mathbb{F}_{SKFF} (\mathbf{F_{mid\downarrow}}, \mathbf{F_{mid\uparrow}})
\end{equation}
where $\mathbb{F}_{SKFF}$ stands for Selective Kernel Feature Fusion (SKFF) module \cite{zamir2022learning} which performs diverse-scale feature integration via dynamic receptive field adjustment.
In the same way, the high-resolution fused feature $\mathbf{F_{high}}$ can be acquired as follows: 
\begin{equation}
\label{eq_hrnet_u2_2}
\mathbf{F_{high\uparrow}}=\mathbb{F}_{u2} (\mathbf{\hat{F}_{mid}})
\end{equation}
\begin{equation}
\label{eq_hrnet_skff_2}
\mathbf{\hat{F}_{high}}=\mathbb{F}_{SKFF} (\mathbf{F_{high}}, \mathbf{F_{high\uparrow}}).
\end{equation}

Finally, the final enhanced image $\mathbf{I^{HR}} $ is generated as follows:
\begin{equation}
\label{eq_hrnet_ct}
\mathbf{I^{HR}}=\mathbb{F}_{tanh}(\mathbb{F}_{conv}(\mathbf{\hat{F}_{high}})).
\end{equation}

From Fig. \ref{fig_intro}(d), it is evident that the haze degradation issue has been effectively addressed by the HR-Net, resulting in a significant enhancement of detail and contrast.

\subsubsection{Hybrid Loss}
As expressed in Eq. \ref{eq_loss_hr}, we will describe the details of $L_{haze}$ and $L_{ctr}^{HR}$. 

\textbf{Structural similarity loss}: The previous work \cite{li2020underwater} has demonstrated that the SSIM \cite{wang2004image} loss can enhance the structural similarity between the enhanced result and the reference image. Therefore, in this work, we also introduce the SSIM loss defined between the final enhanced image $\mathbf{I^{HR}}$ and the referenced image $\mathbf{I^{REF}}$ as follows:
\begin{equation}
\label{eq_lmssim_hr}
{L}_{haze}=1-\frac{1}{H \times W}\sum_{j}{SSIM}(\mathbf{{I}_{j}^{HR}},\mathbf{{I}_{j}^{REF}})
\end{equation}		
where ${SSIM}(\mathbf{{I}_{j}^{HR}},\mathbf{{I}_{j}^{REF}})$ denotes the structure similarity value computed on the $j$-th window.

\textbf{Contrastive loss}: 
We also impose a supplementary contrastive loss in the second stage by \textbf{taking the color corrected image $\mathbf{I^{CC}}$ as negative samples}, yielding a better lower bound than using raw underwater images as negative samples. The contrastive loss is defined as follows:
\begin{equation}
\label{eq_lc_hr}
{L}_{ctr}^{HR}=\frac{1}{s} \sum_{i=1}^{N}w_{i}\,\frac{\left \|\mathbb{E}_{i}({ \mathbf{I^{HR}})-\mathbb{E}_{i}(\mathbf{{I}^{REF}})}\right \|_{1}}{\left \|\mathbb{E}_{i}(\mathbf{I^{HR}})-\mathbb{E}_{i}(\mathbf{I^{CC}})\right \|_{1}}
\end{equation}
where $\mathbb{E}$, $i$, $w_i$ are the same values as in Eq. \ref{eq_lc_cc}, scaling factor $s$ is empirically set to $1$ in this work.

\section{ Experiments}
This section primarily describes the details of the experiments, as well as the analysis of the experimental results. Firstly, we describe the implementation details of the experiments. Then, we clarify the training and testing datasets, comparative methods, and quality assessment metrics. Subsequently, we compare the proposed CCL-Net with the mainstream methods both qualitatively and quantitatively to demonstrate the superiority of our proposed CCL-Net. Finally, we conduct ablation experiments on the key components of the proposed CCL-Net and discuss the limitation of the proposed CCL-Net.

\subsection{Implementation Details}
Our proposed CCL-Net contains two stages: color correction stage and haze removal stage. The training procedure is detailed as follows. First, the color correction stage is trained on $800$ paired raw and reference underwater images in the UIEB dataset \cite{li2019underwater}. Then, the second stage is trained by using the $800$ enhanced images generated in the first stage and their corresponding reference images.

For network training, images are resized to $256 \times 256$. We also perform probabilistic flipping to augment the training data. The training batch size is set to 8, and the Adam optimizer is employed. CC-Net uses a learning rate of 5e-4, while HR-Net employs a learning rate of 1e-3, both with a momentum of 0.5. The training epoch is set to 150. A linearly decreasing learning rate scheduler is adopted from the 75th epoch. The experiments are conducted on PC with a NVIDIA RTX 24G 3090 GPU.

\begin{figure*}[!t]
\centerline{\includegraphics[width=6.2in]{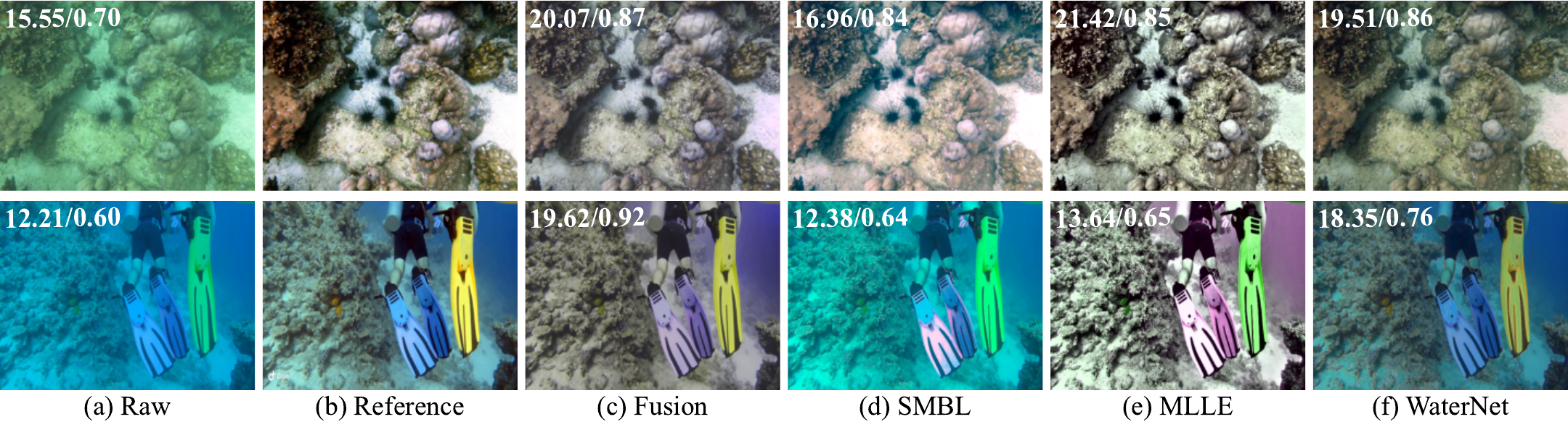}}
\vspace{5pt} 
\centerline{\includegraphics[width=6.2in]{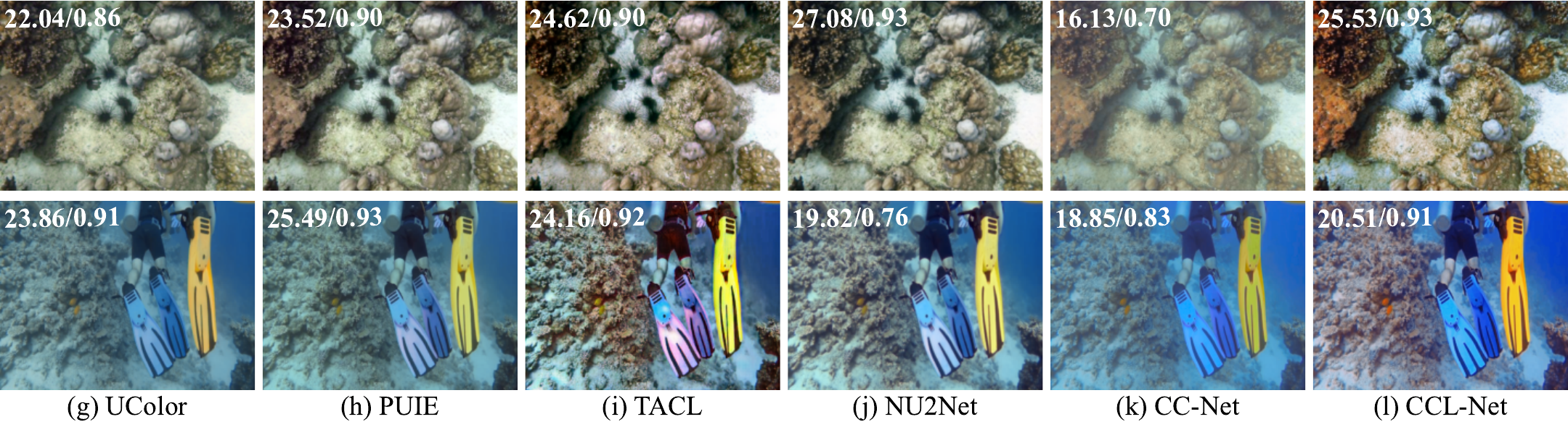}}
\caption{Visual comparisons on underwater images with pronounced greenish and bluish tone from UIEB-T90. The numbers on the top-left corner of each image refer to its PSNR/SSIM values.}
\label{fig_vc_uieb-t90}
\end{figure*}

\subsection{Experiment Settings}
{\bf Test Datasets}. We select the remaining 90 pairs in UIEB\cite{li2019underwater}, referred to as {\bf UIEB-T90}, for testing. Furthermore, to assess the network's generalization performance, we conduct comprehensive test on other four widely-recognized benchmarks, i.e., {\bf UIEB-C60}, EUVP \cite{islam2020fast}, SQUID \cite{berman2020underwater}, and RUIE \cite{liu2020real}, which contain a substantial number of real-world underwater images covering a wide range of degradation types and diverse scenes.
\begin{itemize}
    \item UIEB \cite{li2019underwater} also contains a more challenging subset of 60 images. Some images in this subset have very low brightness, making it hard to be recognized with useful information and the existing UIE methods cannot achieve satisfactory results on {\bf UIEB-C60}. Therefore, the corresponding reference images of {\bf UIEB-C60} are not provided in UIEB. 
    \item EUVP \cite{islam2020fast} is a underwater dataset publicly released by Islam et al. in 2020. The authors employed cycleGAN to learn characteristics of different underwater image types, generating a collection of 12,000 paired underwater original and clear images. We use its test set including 515 images for evaluation, denoted as {\bf EUVP-T515}. 
    \item SQUID \cite{berman2020underwater} is a publicly available real underwater dataset introduced by Berman et al. in 2021, containing 57 pairs of authentic stereoscopic underwater images from four distinct diving locations in Israel. We select 16 representative images, with four chosen from each of the four diving sites, labeled as {\bf SQUID-T16}. 
    \item RUIE \cite{liu2020real} is a real underwater dataset collected from the Yellow Sea in China, publicly released by Liu et al. in 2020. It includes 300 images with color distortion in the UCCS subset, 3630 images with various degradation types in the UIQS subset, and 300 images driven by object detection tasks in the UHTS subset. From these three subsets, we choose 78 representative images, denoted as {\bf RUIE-T78}, which includes 6 images from each of the 13 categories, aiming to cover a wide range of degradation types and diverse scenes.
\end{itemize}

\begin{figure*}[t]
\centerline{\includegraphics[width=6.2in]{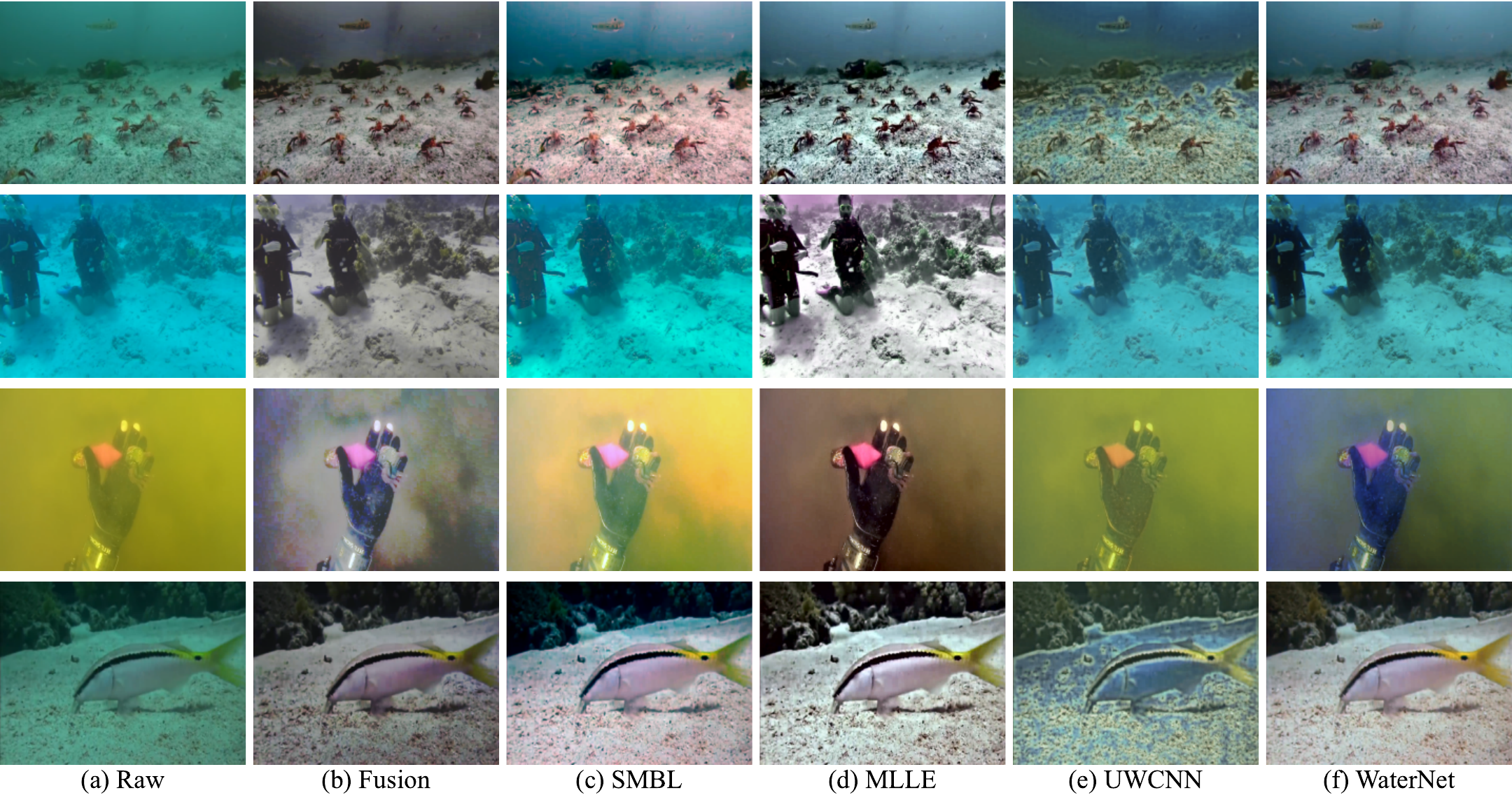}}
\vspace{5pt} 
\centerline{\includegraphics[width=6.2in]{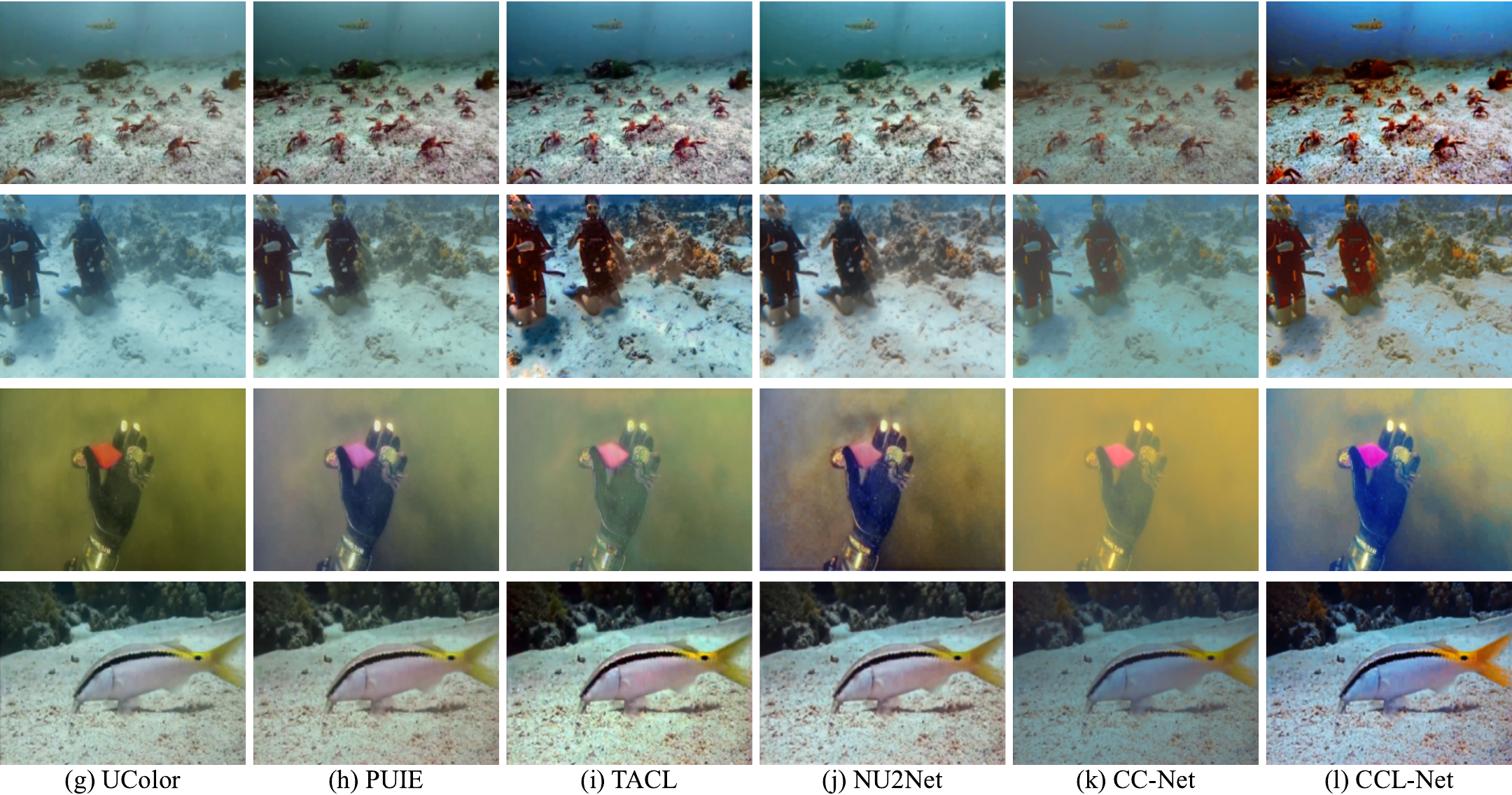}}
\caption{Visual comparisons on challenging underwater images with predominant greenish, bluish, and yellowish tones, as well as low brightness from UIEB-C60.}
\label{fig_vc_uieb-C60}
\end{figure*}

\begin{figure*}[!t]
\centerline{\includegraphics[width=6.4in]{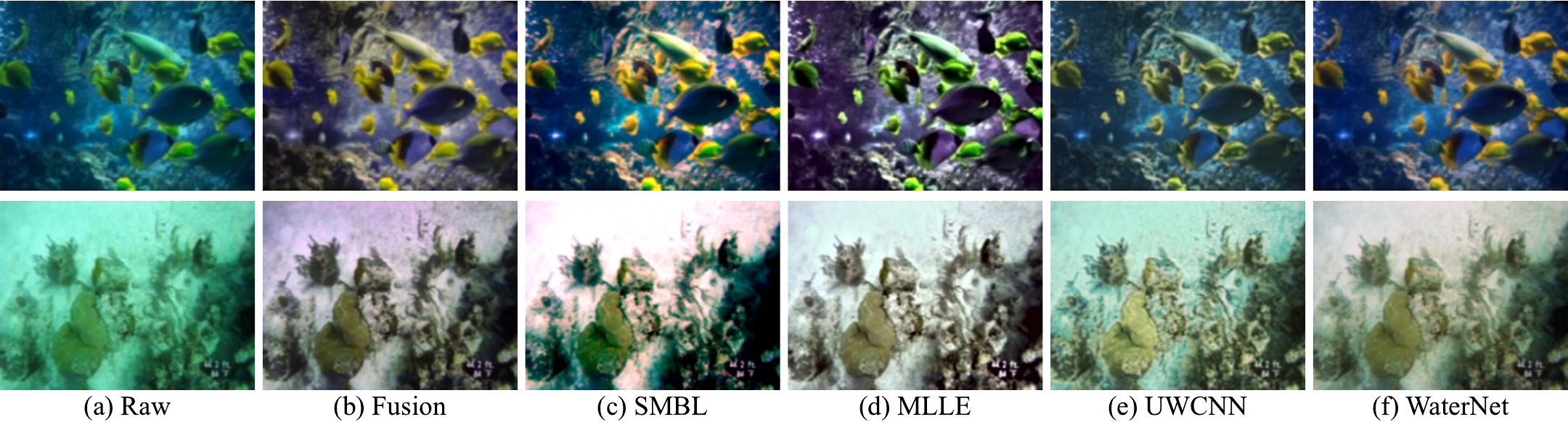}}
\vspace{5pt} 
\centerline{\includegraphics[width=6.4in]{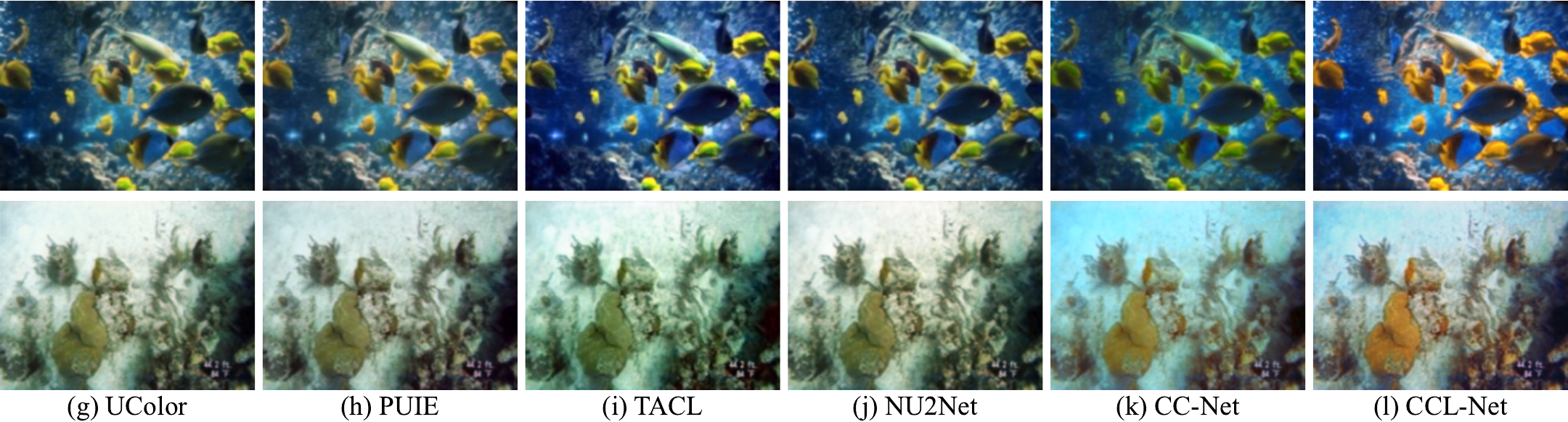}}
\caption{Visual comparisons on underwater images with greenish color deviation and low contrast from EUVP-T515.}
\label{fig_vc_euvp-t515}
\end{figure*}

\begin{figure*}[!t]
\centerline{\includegraphics[width=6.2in]{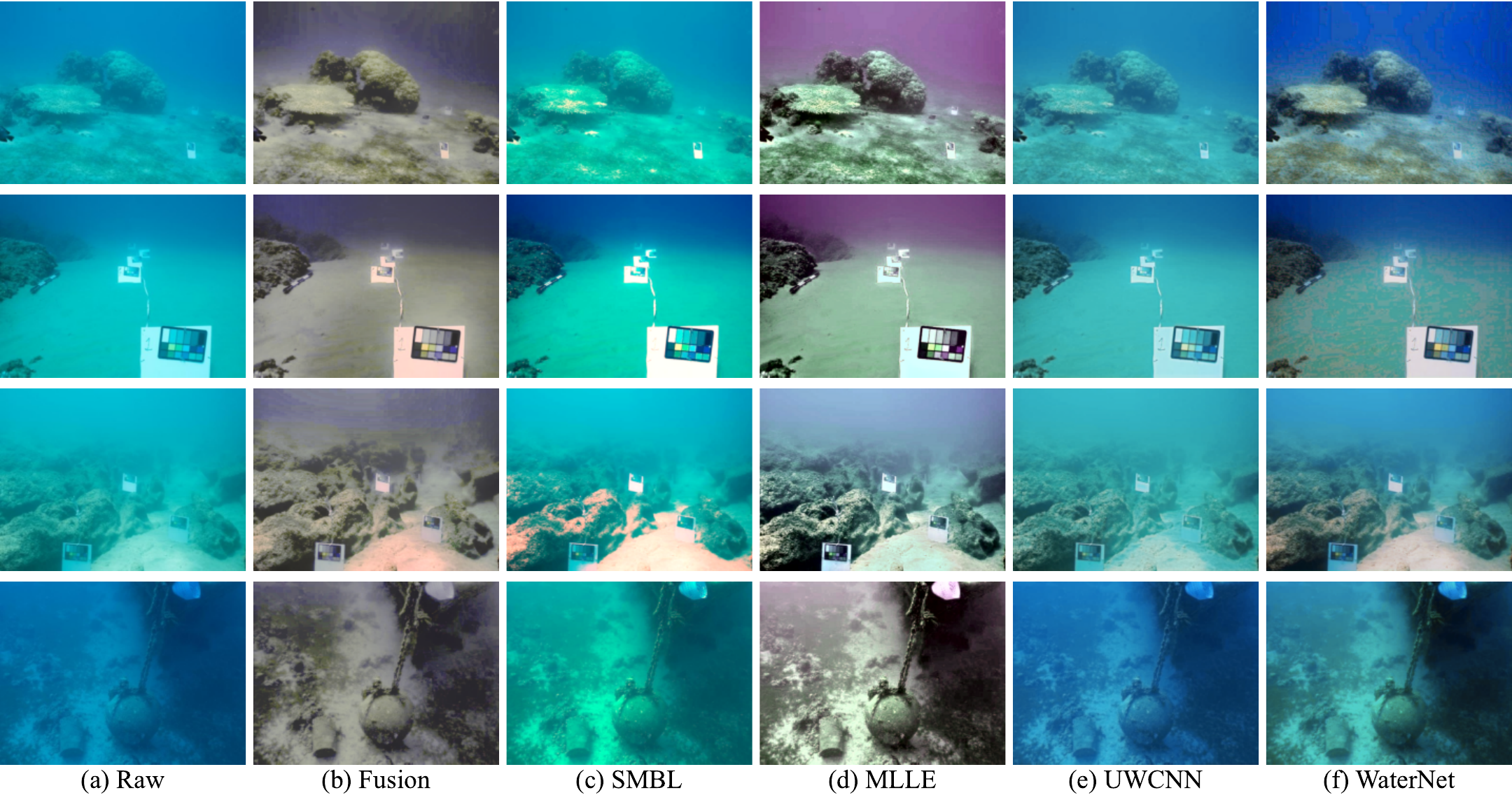}}
\vspace{5pt} 
\centerline{\includegraphics[width=6.2in]{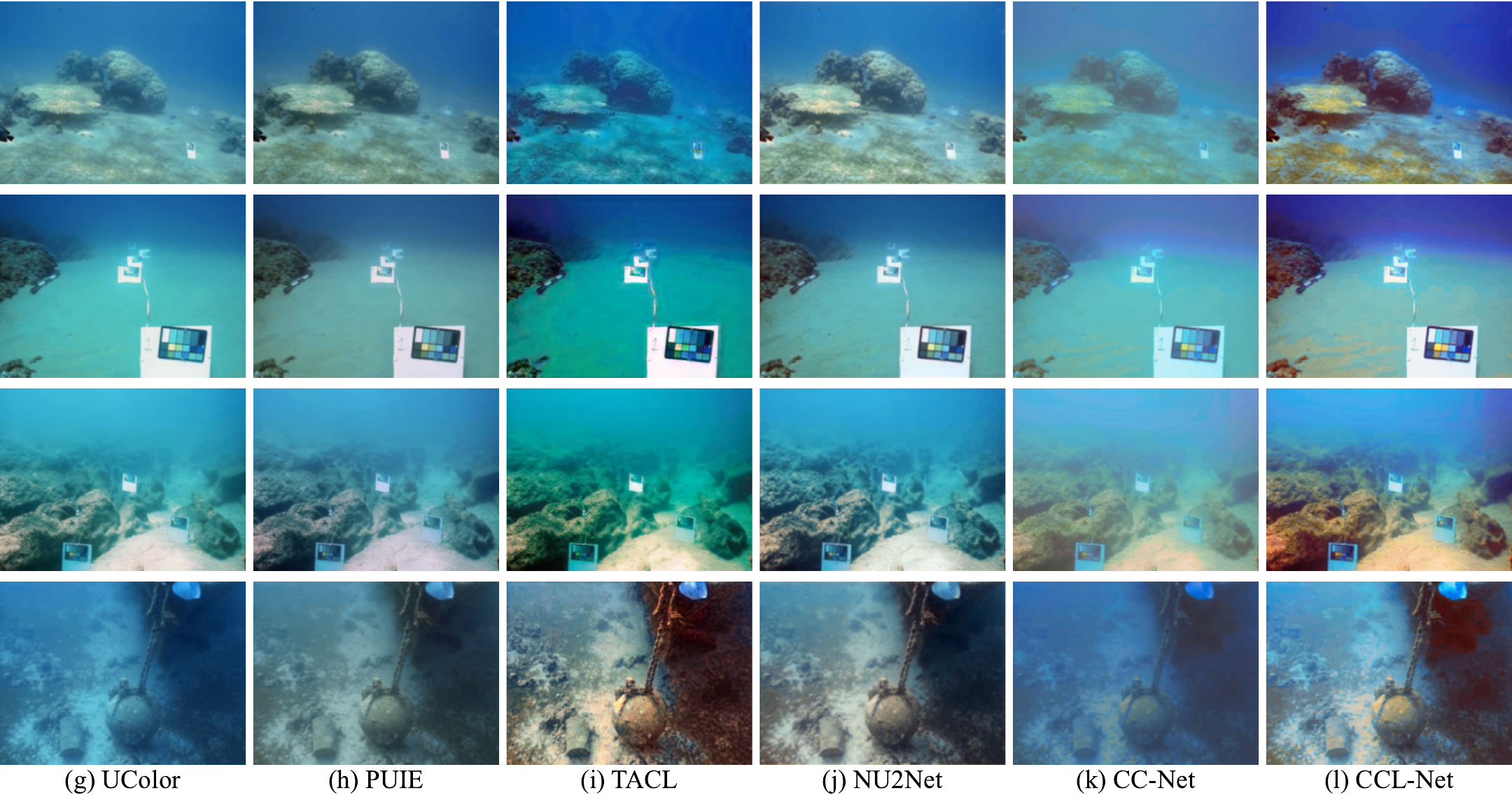}}
\caption{Visual comparisons on underwater images with different levels of bluish color deviation from SQUID-T16.}
\label{fig_vc_squid-t16}
\end{figure*}

{\bf Compared Methods}. We conduct comparisons with several state-of-the-art (SOTA) methods, including three traditional methods and six deep learning methods. The three traditional methods are Fusion \cite{ancuti2017color}, SMBL \cite{song2020enhancement}, and MLLE \cite{zhang2022underwater}. The six deep learning methods are UWCNN \cite{li2020underwater}, WaterNet\cite{li2019underwater}, UColor\cite{li2021underwater}, PUIE\cite{fu2022uncertainty}, TACL\cite{liu2022twin}, and NU2Net\cite{guo2023underwater}. Note that all of these deep learning methods are built with the single-stage framework. For Fusion \cite{ancuti2017color}, as the author does not release the source code, we utilize an open-source implementation found online\footnote{https://github.com/fergaletto/Color-Balance-and-fusion-for-underwater-image-enhancement.-.}. For the implementation of SMBL\cite{song2020enhancement} and MLLE \cite{zhang2022underwater}, we employ the source code released by the authors. For UWCNN\cite{li2020underwater}, we utilize an open-source PyTorch implementation available online\footnote{https://github.com/BIGWangYuDong/UWEnhancement} and retrain the network using the same training dataset as our network. For WaterNet\cite{li2019underwater}, UColor\cite{li2021underwater}, PUIE\cite{fu2022uncertainty}, TACL\cite{liu2022twin}, and NU2Net\cite{guo2023underwater}, we conduct experiments using the authors' publicly available model source codes and pre-trained models.

{\bf Evaluation Metrics}. For UIEB-T90, we conduct a comprehensive evaluation using both reference-based metrics and non-reference metrics. This allows us to assess the strength and weakness of different methods from various perspectives. As for the other testing datasets, due to the reference images not being real or the absence of reference images, accurate assessment using reference-based metrics is not feasible. Therefore, we only employ non-reference metrics for performance evaluation. The reference-based metrics include PSNR and SSIM \cite{wang2004image}. Higher PSNR and SSIM values indicate that the enhanced images are closer to the reference images. The non-reference metrics consist of UIQM \cite{panetta2015human} and UCIQE \cite{yang2015underwater}, specifically designed for underwater image quality assessment. Higher UIQM and UCIQE values indicate better visual quality perceived by humans.

\subsection{Visual Comparisons}
In this subsection, we provide the visual results of different UIE methods on some representative examples with different underwater scenes and degradation types selected from the above-mentioned test datasets. In the following, the results on each test dataset are presented and discussed with details.

{\bf UIEB-T90}: Two representative images with pronounced green and blue color deviations are selected from UIEB-T90 for comparison, as shown in Fig. \ref{fig_vc_uieb-t90}(a). Due to the influence of serious color deviations, the objects in these images cannot be easily perceived, resulting in a relatively lower visual quality. For instance, the black sea urchin and the red fish in Fig. \ref{fig_vc_uieb-t90}(a). The color correction and haze removal results produced by our two-stage framework based method are shown in Fig. \ref{fig_vc_uieb-t90} (k) and (l), respectively. From Fig. \ref{fig_vc_uieb-t90} (k), it can be observed that the green and blue deviations have been corrected by our initial CC-Net. The original color of objects are mostly restored. However, the results from the CC-Net in Fig. \ref{fig_vc_uieb-t90} (k) still exhibit a hazy appearance. With the help of HR-Net, the clarity of object details and color saturation have been significantly improved, as shown in Fig. \ref{fig_vc_uieb-t90} (l). Furthermore, the contrast between objects is also enhanced, leading to greater distinction among different objects. Overall, the enhanced images generated by the two-stage framework-based method can achieve an impressive improvement in visual quality compared to the original images, thus demonstrating the effectiveness of our proposed CCL-Net. Fig. \ref{fig_vc_uieb-t90} (c)-(j) show the enhanced results of recent SOTA methods. Most methods manage to address color distortion and haze to a certain extent, achieving better visual quality than the original images. However, three issues still persist. First, some methods may introduce additional color deviations beyond the common greenish and blueish appearances. For example, in the $2$nd row of Fig. \ref{fig_vc_uieb-t90}, the enhanced image by SMBL exhibits an overall green bias, with fins turning purple as well as TACL. Additionally, the background water appears purplish and pink in the results of Fusion and MLLE, and two blue and yellow fins in the result of MLLE are changed to pink-purple and green, respectively. Second, haze effect may still present, and object details are lacking of clarity. In the $4$th row of Fig. \ref{fig_vc_uieb-t90}, slight haze is still noticeable in the enhanced images obtained by WaterNet, UColor, and PUIE. Third, object color saturation is insufficient, and inter-object contrast is low. For blue-deviated images, all compared methods lack full saturation of the blue color in the background sea. Moreover, due to insufficient contrast between fish and rocks, the presence of the orange fish is not easily perceivable in the results of all compared methods, especially Fusion, SMBL, and MLLE. For green-deviated images, the color saturation of object detail and inter-object contrast in all compared methods need to be improved. By contrast, the results of our CCL-Net in Fig. \ref{fig_vc_uieb-t90} (k) and (l) effectively addresses color distortion and haze issues, resulting in enhanced images that are more natural, detailed, saturated in color, and have higher contrast. The $1$st image also attains the $2$nd highest PSNR score and the highest SSIM score. Obviously, the compared results demonstrate the superiority of our proposed CCL-Net.

{\bf UIEB-C60}: For the challenging scenarios in UIEB-C60, we select four representative images with greenish, blueish, yellowish tones, and low brightness as the raw images for comparison, as shown in Fig. \ref{fig_vc_uieb-C60} (a). For these four images, none of the compared methods obtain satisfactory results. Firstly, some compared methods fail to effectively address color deviation, as evident in the $2$nd with SMBL, UWCNN, and WaterNet, where the bluish tone is not adequately eliminated in their enhanced results. Furthermore, some compared methods cannot obtain ideal dehazing results. For instance, in the $6$th row, the enhanced results of UColor and PUIE are still with haze effect. Similarly, in the $7$th row, the enhanced result of TACL also exhibits slight haze effect. Moreover, the enhanced results of some compared methods suffer from low color saturation and contrast, e.g., Fusion and MLLE. By contrast, our CCL-Net effectively resolves the aforementioned issues, generating enhanced results with minimal color deviation and weakest haze effect, as shown in Fig. \ref{fig_vc_uieb-C60} (l). 

{\bf EUVP-T515}: Let's further analyze the comparison results for EUVP-T515. As shown in the $1$st row of Fig. \ref{fig_vc_euvp-t515}, some conventional methods, e.g., Fusion, SMBL, and MLLE, introduce extra color artifacts, resulting in purple background and green fish. For the $2$nd image, the results of all compared methods effectively eliminate the green color deviation. However, the original colors of the objects are not well restored, and the perceived image contrast is also limited. By contrast, as shown in columns (k) and (l) of Fig. \ref{fig_vc_euvp-t515}, our method can effectively recover the original color and remove haze. From Fig. \ref{fig_vc_euvp-t515} column (l), it can be observed that our CCL-Net not only restores the natural color of fish and background in the $1$st image, but also improves the color saturation and contrast of the brown object in the $2$nd image.

{\bf SQUID-T16}: For the visual comparison results on different levels of blue-deviated underwater images contained in SQUID-T16, as shown in Fig. \ref{fig_vc_squid-t16}, it is evident that all compared methods struggle to achieve visually pleasing effect on this dataset. Some compared methods fail to correct color distortion. For instance, SMBL and MLLE introduces greenish and purplish color deviations, respectively, while UWCNN hardly achieves any color correction effect. In the case of TACL, the $1$st image exhibits a bluish tone, and the $2$nd and $3$rd images tend to have a greenish tone. Some methods exhibit limited haze removal effect. For example, UColor and PUIE obtain moderate color correction effect, yet a slight haze effect still exists. The enhanced results of most compared methods present low color saturation and contrast. For instance, Fusion effectively removes blue deviation, but there is little color discrepancy among different objects. By contrast, our CCL-Net achieves the desired visual effect. Specifically, color distortion is effectively addressed yet haze effect still exists after passing through the CC-Net, as shown in Fig. \ref{fig_vc_squid-t16} (k). As seen in Fig. \ref{fig_vc_squid-t16} (l), the CCL-Net efficiently removes haze and enhances color saturation and contrast, resulting in a visually remarkable effect.

\begin{figure*}[!t]
\centerline{\includegraphics[width=6.2in]{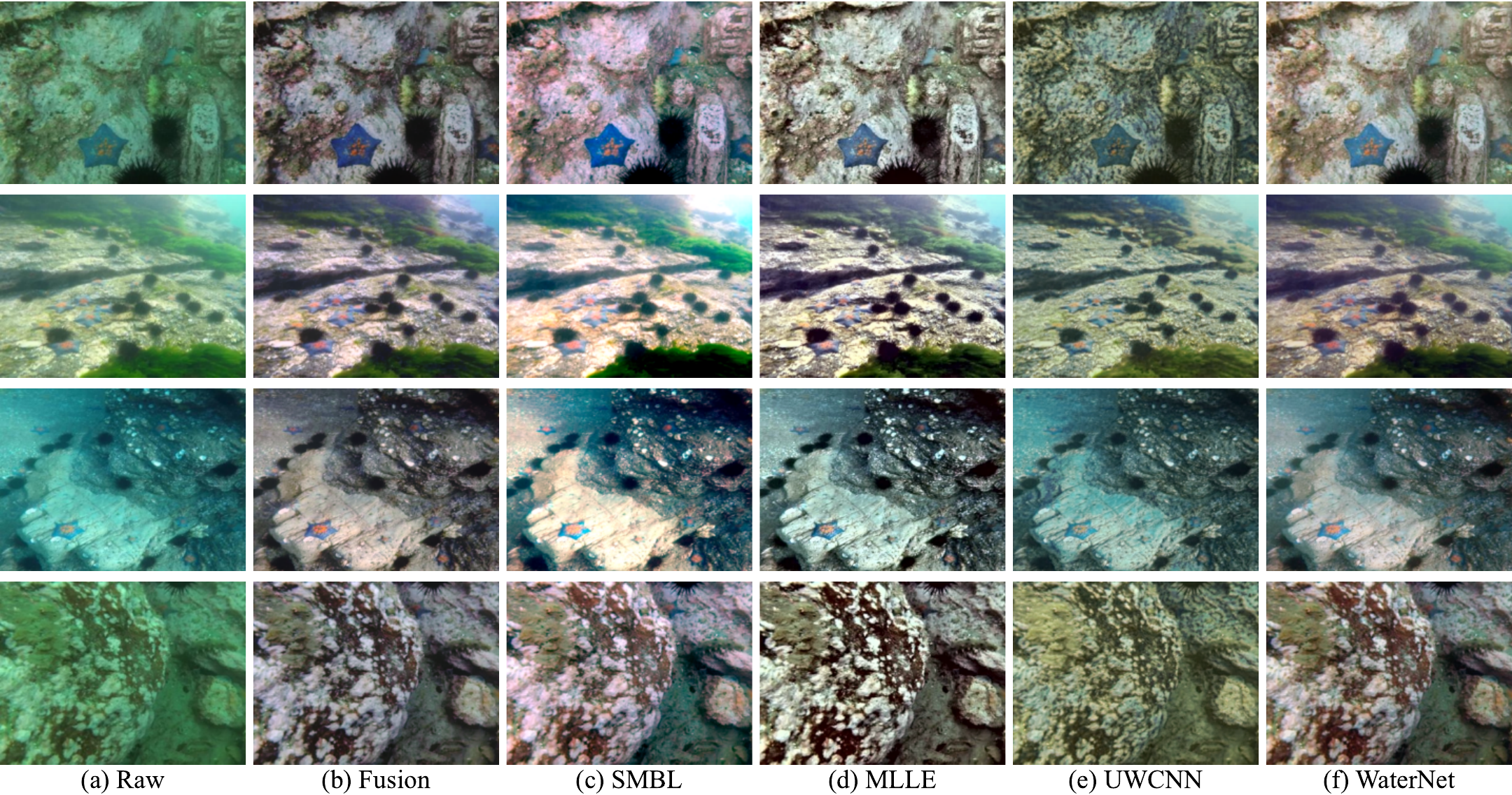}}
\vspace{5pt} 
\centerline{\includegraphics[width=6.2in]{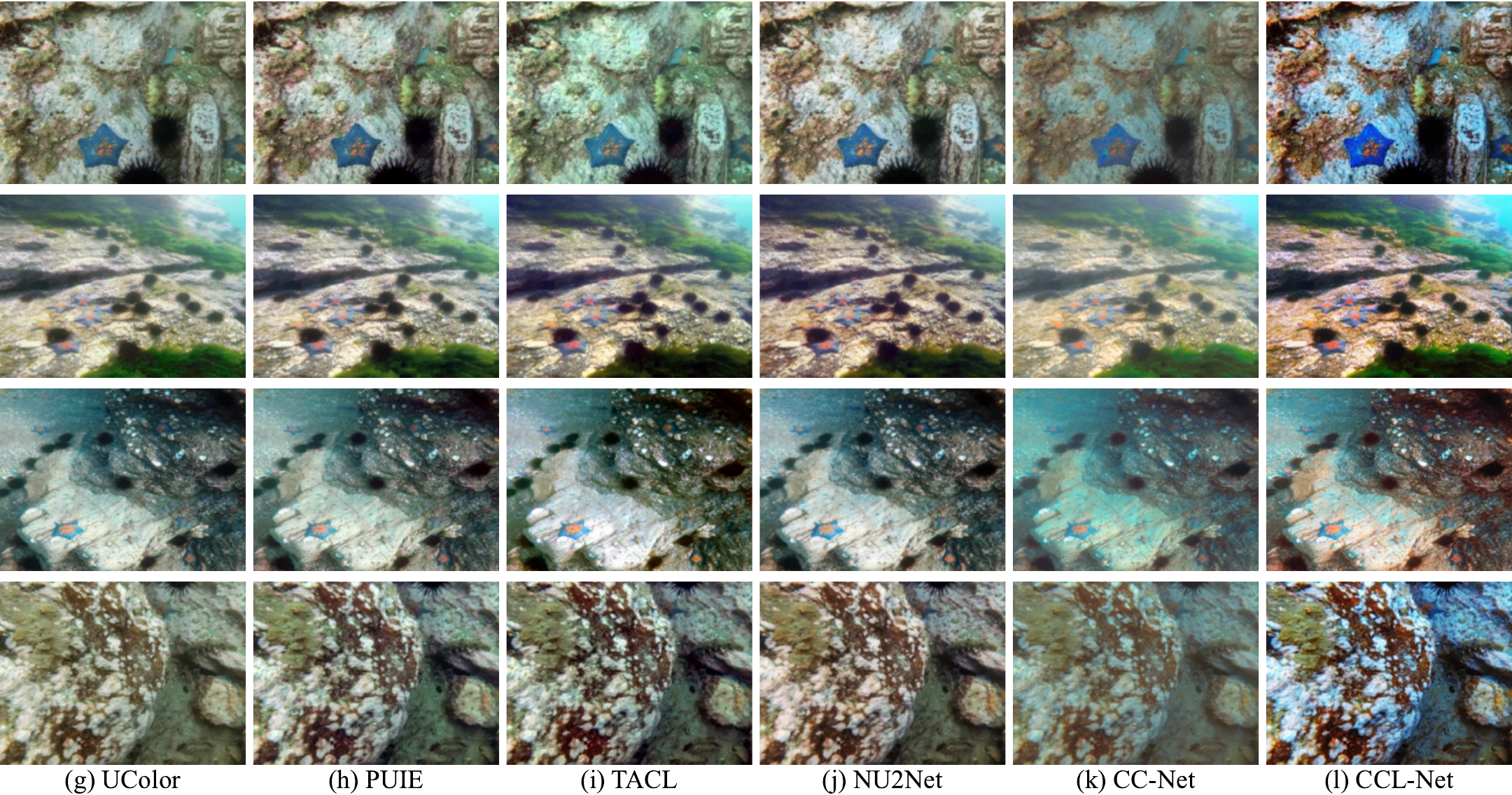}}
\caption{Visual comparisons on underwater images with four degradation types including hazy, yellowish, bluish, and greenish color deviations from RUIE-T78.}
\label{fig_vc_ruie-t78}
\end{figure*}

{\bf RUIE-T78}: For comparisons on RUIE-T78, we select four representative images with hazy effect, yellowish, bluish, and greenish tones as raw images, as shown in Fig. \ref{fig_vc_ruie-t78} (a). 
Most compared methods achieve satisfactory visual effect. However, some methods still exhibit a slight haze effect and low brightness. For the enhanced results of WaterNet and UColor on the $1$st raw image, the presence of thin haze leads to the low color saturation of image. For the enhanced results of Fusion and UWCNN on the $1$st and $4$th raw images, the overall brightness is low. By contrast, our proposed CCL-Net effectively addresses these issues. As presented in Fig. \ref{fig_vc_ruie-t78} (l), CCL-Net performs effective haze removal and brightness adjustment, resulting the enhanced images with clear structure, high color saturation and contrast. 

The above comprehensive comparative experiments on multiple real underwater datasets fully demonstrate the visual effectiveness and superiority of our method, confirming the rationality of our method's design.

\subsection{Quantitative Comparisons}
To further demonstrate the effectiveness and superiority of the proposed CCL-Net from a quantitative metric perspective, a comparison of different methods on the UIEB-T90 dataset is presented in Table \ref{tbl_qc_uieb-t90}. The scores in the table represent the mean results for all 90 images. Our method shows competitive performance in terms of both reference-based and non-reference metrics. Specifically, our method outperforms traditional methods in terms of PSNR and SSIM scores. This suggests that traditional methods struggle to handle complex real underwater scenes and exhibit weaker generalization capability. Although our method does not achieve optimal PSNR and SSIM scores, it may be due to the inadequate consideration of underwater image characteristics by PSNR and SSIM, leading to unreliable results. As observed from the visual comparison in Fig. \ref{fig_vc_uieb-t90} ($2$nd image), our method achieves the best human visual quality, while the PSNR and SSIM scores are not the highest. Therefore, we focus more on the non-reference metrics. For UIQM and UCIQE, our method scores the highest in UIQM and the second highest in UCIQE (only slightly lower than MLLE by 1\%) among all compared methods, resulting in the best performance among all deep learning methods. 

\begin{table}[!t]
    \centering
    \caption{Quantitative comparison on UIEB-T90 in terms of PSNR, SSIM, UIQM and UCIQE. The top three scores are marked in  {\color[HTML]{FF0000} red}, {\color[HTML]{00B050} green}, and {\color[HTML]{00B0F0} blue}.}
    \label{tbl_qc_uieb-t90}
    \begin{tabular}{l||cccc}
\hline
\textbf{Methods}          & \textbf{PSNR↑}                & \textbf{SSIM↑}               & \textbf{UIQM↑}               & \textbf{UCIQE↑}              \\ \hline \hline
Raw                    & 16.134                        & 0.748                        & 2.346                        & 0.362                        \\
Fusion(TIP'17)\cite{ancuti2017color}   & 18.033                        & 0.861                        & 2.684                        & 0.406                        \\
SMBL(TB'20)\cite{song2020enhancement}      & 16.513                        & 0.781                        & 2.167                        & {\color[HTML]{00B0F0} 0.455} \\
MLLE(TIP'22)\cite{zhang2022underwater}     & 18.727                        & 0.790                        & 2.305                        & {\color[HTML]{FF0000} 0.468} \\
UWCNN(PR'20)\cite{li2020underwater}      & 18.147                        & 0.847                        & 2.878                        & 0.357                        \\
WaterNet(TIP'19)\cite{li2019underwater} & 19.914                        & 0.859                        & 2.846                        & 0.410                        \\
UColor(TIP'21)\cite{li2021underwater}   & 20.933                        & {\color[HTML]{00B050} 0.869} & 2.869                        & 0.391                        \\
PUIE(ECCV'22)\cite{fu2022uncertainty} & {\color[HTML]{00B050} 22.023} & {\color[HTML]{FF0000} 0.893} & 2.849                        & 0.396                        \\
TACL(TIP'22)\cite{liu2022twin}     & {\color[HTML]{00B0F0} 22.735} & 0.864                        & {\color[HTML]{00B050} 3.016} & 0.445                        \\
NU2Net(AAAI'23)\cite{guo2023underwater}  & {\color[HTML]{FF0000} 22.820} & {\color[HTML]{FF0000} 0.893} & {\color[HTML]{00B0F0} 2.902} & 0.422                        \\ \hline
CCL-Net                     & 20.181                        & {\color[HTML]{00B0F0} 0.866} & {\color[HTML]{FF0000} 3.021} & {\color[HTML]{00B050} 0.464}   \\ \hline
\end{tabular}
\end{table}

\begin{table*}[!t]
\centering
\caption{Quantitative comparison on UIEB-C60, EUVP-T515, SQUID-T16 and RUIE-T78 in terms of UIQM and UCIQE. The top three scores are marked in {\color[HTML]{FF0000} red}, {\color[HTML]{00B050} green}, and {\color[HTML]{00B0F0} blue}.}
\label{tbl_qc_nr}
\renewcommand{\arraystretch}{1.1} % 设置行距为原始行高的1.5倍

\begin{tabular}{l||cc|cc|cc|cc}
\hline
\multicolumn{1}{c||}{}                                  & \multicolumn{2}{c|}{\textbf{UIEB-C60}}                                    & \multicolumn{2}{c|}{\textbf{EUVP-T515}}                                   & \multicolumn{2}{c|}{\textbf{SQUID-T16}}                                   & \multicolumn{2}{c}{\textbf{RUIE-T78}}                                    \\ \cline{2-9} 
\multicolumn{1}{c||}{\multirow{-2}{*}{\textbf{Methods}}} & \multicolumn{1}{l}{\textbf{UIQM↑}} & \multicolumn{1}{l|}{\textbf{UCIQE↑}} & \multicolumn{1}{l}{\textbf{UIQM↑}} & \multicolumn{1}{l|}{\textbf{UCIQE↑}} & \multicolumn{1}{l}{\textbf{UIQM↑}} & \multicolumn{1}{l|}{\textbf{UCIQE↑}} & \multicolumn{1}{l}{\textbf{UIQM↑}} & \multicolumn{1}{l}{\textbf{UCIQE↑}} \\ \hline \hline
Raw                                                  & 1.856                              & 0.359                                & 2.217                              & 0.417                                & 0.655                              & 0.393                                & 2.437                              & 0.321                               \\
Fusion(TIP'17)\cite{ancuti2017color}                                 & 2.163                              & 0.378                                & 2.636                              & 0.430                                & 1.860                              & 0.342                                & 2.772                              & 0.366                               \\
SMBL(TB'20)\cite{song2020enhancement}                                    & 1.724                              & {\color[HTML]{00B050} 0.439}         & 1.857                              & {\color[HTML]{FF0000} 0.513}         & 0.762                              & {\color[HTML]{00B050} 0.452}         & 2.459                              & {\color[HTML]{00B0F0} 0.431}        \\
MLLE(TIP'22)\cite{zhang2022underwater}                                   & 1.956                              & {\color[HTML]{FF0000} 0.464}         & 2.354                              & {\color[HTML]{00B050} 0.461}         & 2.151                              & {\color[HTML]{00B0F0} 0.431}         & 2.798                              & {\color[HTML]{00B050} 0.441}        \\ 
UWCNN(PR'19)\cite{li2020underwater}                                    & 2.433                              & 0.340                                & {\color[HTML]{00B0F0} 2.822}       & 0.369                                & 1.804                              & 0.361                                & 3.053                              & 0.314                               \\
WaterNet(TIP'20)\cite{li2019underwater}                               & 2.468                              & 0.402                                & 2.680                              & 0.412                                & {\color[HTML]{00B050} 2.206}       & 0.402                                & {\color[HTML]{00B0F0} 3.115}       & 0.403                               \\
UColor(TIP'21)\cite{li2021underwater}                                 & 2.411                              & 0.371                                & 2.749                              & 0.404                                & 1.987                              & 0.369                                & 3.006                              & 0.345                               \\
PUIE(ECCV'22)\cite{fu2022uncertainty}                               & 2.379                              & 0.375                                & 2.748                              & 0.407                                & 2.090                              & 0.334                                & 2.989                              & 0.379                               \\
TACL(TIP'22)\cite{liu2022twin}                                   & {\color[HTML]{FF0000} 2.854}       & 0.424                                & {\color[HTML]{00B050} 2.837}       & 0.435                                & 2.202                              & {\color[HTML]{FF0000} 0.470}         & {\color[HTML]{FF0000} 3.237}       & 0.422                               \\
NU2Net(AAAI'23)\cite{guo2023underwater}                                & {\color[HTML]{00B0F0} 2.508}       & 0.409                                & 2.767                              & 0.422                                & {\color[HTML]{00B0F0} 2.205}       & 0.386                                & 3.061                              & 0.389                               \\ \hline
CCL-Net                                                   & {\color[HTML]{00B050} 2.622}       & {\color[HTML]{00B0F0} 0.434}         & {\color[HTML]{FF0000} 2.936}       & {\color[HTML]{00B0F0} 0.456}         & {\color[HTML]{FF0000} 2.350}       & 0.404                                & {\color[HTML]{00B050} 3.168}       & {\color[HTML]{FF0000} 0.447}        \\ \hline
\end{tabular}
\end{table*}

We also report the average UIQM and UCIQE scores of different methods on the other four testing datasets, i.e., UIEB-C60, EUVP-T515, SQUID-T16, and RUIE-T78, as shown in Table \ref{tbl_qc_nr}. In terms of UIQM scores, our method demonstrates remarkable competitiveness. More precisely, our method achieves the highest scores on the EUVP-T515 and SQUID-T16 datasets, and the second highest scores on the UIEB-C60 and RUIE-T78 datasets. Regarding UCIQE score, our method achieves highly competitive scores. More specifically, it ranks $1$st on the RUIE-T78 dataset. On the UIEB-C60, EUVP-T515, and SQUID-T16 datasets, it is always among the top-four. This indicates the superior effectiveness, robustness, and generalization ability of our method compared to most of the competing methods. However, two points deserve attention. First, although traditional methods SMBL and MLLE achieve higher UCIQE scores than our method, the visual comparisons reveal that these two methods perform poorly on the UIEB-C60, EUVP-T515, and SQUID-T16 datasets. As reported in previous works \cite{li2021underwater}\cite{kang2022perception}, the UCIQE metric cannot truly measure human perception of underwater enhanced image quality. Therefore, we claim that the best way to use non-reference metrics for quantitative comparison is to combine the non-reference metric results and subjective visual judgments. On considering these two aspects, our method possesses greater reliability. Second, our proposed two-stage framework method outperforms all single-stage framework deep learning methods except TACL in terms of both UIQM and UCIQE scores, which is basically consistent with the above visual comparison results. These comprehensive comparisons from multiple perspectives strongly demonstrate the advantage of our proposed two-stage framework over current mainstream single-stage framework in deep learning-based UIE, providing a novel exploration mode for future research in this direction.

\subsection{Ablation Study}
To validate the effect of several key components of our proposed CCL-Net, we conduct ablation experiments with the following five models: 1) \textbf{Full}: the complete model proposed in this paper; 2) \textbf{w/o CC-Net}: remove CC-Net while retaining HR-Net, 3) \textbf{w/o HR-Net}: remove HR-Net while retaining CC-Net; 4) \textbf{w/o CL}: without using the contrastive loss, 5) \textbf{RAN}: using Raw underwater image As Negative (RAN) samples to build the contrastive loss in HR-Net.

\begin{table}[!t]
\centering
    \caption{Quantitative results of the ablation models on UIEB-T90 and UIEB-C60. The top scores are marked in {\color[HTML]{FF0000} red}.}
    \label{tab3}
\renewcommand{\arraystretch}{1.1} 
\label{tbl_as}
\begin{tabular}{l||c@{\hspace{3pt}}c@{\hspace{3pt}}c@{\hspace{3pt}}c|@{\hspace{3pt}}c@{\hspace{3pt}}c}
\hline
\multicolumn{1}{c||}{}                                    & \multicolumn{4}{c|@{\hspace{3pt}}}{\textbf{UIEB-T90}}                                                                                     & \multicolumn{2}{c}{\textbf{UIEB-C60}}                       \\ \cline{2-7} 
\multicolumn{1}{c||}{\multirow{-2}{*}{\textbf{\makecell[l]{Ablation \\Models}}}} & \textbf{PSNR↑}                & \textbf{SSIM↑}               & \textbf{UIQM↑}               & \textbf{UCIQE↑}              & \textbf{UIQM↑}               & \textbf{UCIQE↑}         \\ \hline \hline
Raw                                                    & 16.134                        & 0.748                        & 2.346                        & 0.362                        & 1.856                        & 0.359                        \\
Full                                                     & 20.181                        & 0.866                        & {\color[HTML]{FF0000} 3.021} & {\color[HTML]{FF0000} 0.464} & {\color[HTML]{FF0000} 2.622} & {\color[HTML]{FF0000} 0.434} \\
w/o CC-Net                                               & {\color[HTML]{FF0000} 22.006} & {\color[HTML]{FF0000} 0.905} & 2.986                        & 0.434                        & 2.499                        & 0.402                        \\
w/o HR-Net                                               & 17.167                        & 0.755                        & 2.546                        & 0.372                        & 2.048                        & 0.353                        \\
w/o CL                                                   & 21.724                        & 0.891                        & 2.940                        & 0.410                        & 2.452                        & 0.378                        \\
RAN                                               & 21.845                        & 0.890                        & 2.939                        & 0.414                        & 2.452                        & 0.384                        \\ \hline
\end{tabular}
\end{table}

\begin{figure}[!t]
\centerline{\includegraphics[width=3.2in]{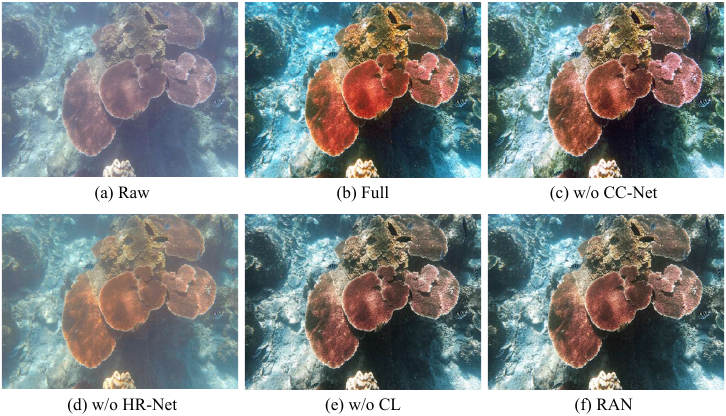}}
\caption{Ablation study of the contributions of each key component. Our full model obtains the best visual quality.}
\label{fig_as_uieb-t90}
\end{figure}

\begin{figure}[!t]
\centerline{\includegraphics[width=3.2in]{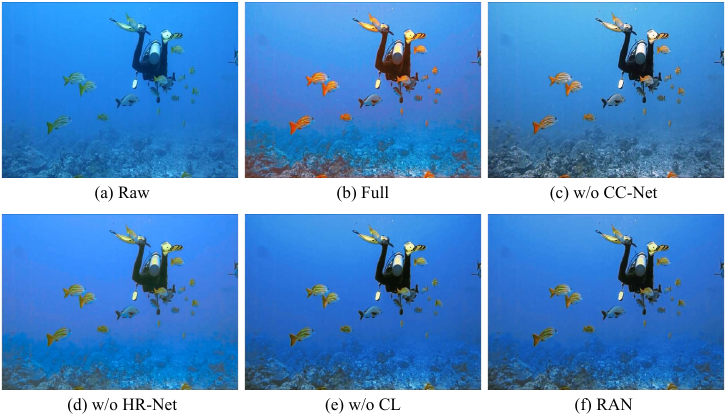}}
\caption{Ablation study of the contributions of each key component. Our full model obtains the best visual quality.}
\label{fig_as_uieb-c60}
\end{figure}

Table \ref{tbl_as} presents the quantitative comparison results of the above ablation models  on the UIEB-T90 and UIEB-C60 datasets. Fig. \ref{fig_as_uieb-t90} and Fig. \ref{fig_as_uieb-c60} depict the visual comparison results on the UIEB-T90 and UIEB-C60 datasets, respectively. Based on these comparison results, the following conclusions can be drawn. \textbf{First}, despite the comparative models outperform the full model in reference-based metrics (PSNR and SSIM), the visual comparisons and non-reference metrics consistently favor the full model, demonstrating the effectiveness of the proposed two-stage framework coupled with cascaded contrastive loss. \textbf{Second}, as evident from the comparison results of w/o CC-Net in Fig. \ref{fig_as_uieb-t90} and Fig. \ref{fig_as_uieb-c60}, although the single-stage HR-Net indeed improves the visual quality to a certain extent, the results still have much room for further improvement in terms of both color saturation and contrast compared to the full model. This implies that the single-stage HR-Net struggles to effectively address both color distortion and haze issues simultaneously. From the comparison results of w/o HR-Net in Fig. \ref{fig_as_uieb-t90}, it is evident that, while color distortion is corrected, the haze effect cannot be well addressed. In contrast, our full model which first performs color correction and then followed by haze removal produces superior visually enhanced image and performance. This highlights the significance of designing a two-stage framework tailored to domain-specific issues. \textbf{Third}, the results of the other ablation models, i.e., w/o CL and RAN, demonstrate that incorporating contrastive loss and higher-quality than the original raw images as negative samples in building the contrastive loss not only improve the non-reference metric performance but also improves color saturation and contrast of the enhanced results.

\begin{table}[!t]
\centering
    \caption{Comparison of different UIE models in terms of FLOPs (G), parameters (M) and running time (Sec). The top three values are marked in {\color[HTML]{FF0000} red}, {\color[HTML]{00B050} green}, and {\color[HTML]{00B0F0} blue}, respectively.}
    \label{tab4}
\renewcommand{\arraystretch}{1.1} % 设置行距为原始行高的1.5倍
\label{tbl_fpr}
\begin{tabular}{l||c|c|c}
\hline
\textbf{Methods}          & \textbf{\begin{tabular}[c]{@{}c@{}}FLOPs \\ (G) ↓\end{tabular}}          & \textbf{\begin{tabular}[c]{@{}c@{}}Parameters \\ (M)↓\end{tabular}}    & \textbf{\begin{tabular}[c]{@{}c@{}}Running \\ Time (s)↓\end{tabular}}  \\ \hline \hline
Fusion(TIP'18)\cite{ancuti2017color}   & -                             & -                           & 0.81                        \\ 
SMBL(TB'20)\cite{song2020enhancement}       & -                             & -                           & 6.78                        \\ 
MLLE(TIP'22)\cite{zhang2022underwater}      & -                             & -                           & 0.19                        \\ 
UWCNN(PR'20)\cite{li2020underwater}      & {\color[HTML]{FF0000} 11.36}  & {\color[HTML]{FF0000} 0.04} & {\color[HTML]{FF0000} 0.01} \\ 
WaterNet(TIP'19)\cite{li2019underwater}  & 310.82                        & 1.09                        & {\color[HTML]{00B0F0} 0.04} \\ 
UColor(TIP'21)\cite{li2021underwater}    & 4345.28                       & 105.51                      & 0.36                        \\ 
PUIE(ECCV'22)\cite{fu2022uncertainty}  & 2073.2                        & {\color[HTML]{00B0F0} 0.83} & 0.40                         \\ 
TACL(TIP'22)\cite{liu2022twin}      & {\color[HTML]{00B0F0} 247.46} & 11.37                       & {\color[HTML]{00B050} 0.03} \\ 
NU2Net(AAAI'23)\cite{guo2023underwater}   & {\color[HTML]{00B050} 46.33}  & 3.15                        & {\color[HTML]{FF0000} 0.01} \\ \hline
CCL-Net                   & 470.62                        & {\color[HTML]{00B050} 0.55} & 0.06                        \\ \hline
\end{tabular}
\end{table}

\subsection{Discussion}
\subsubsection{Model Size and Running Time}
We compare the running time of all UIE methods, as well as the model parameters and FLOPs of all deep UIE methods. The image with a resolution of 620 $\times$ 460 $\times$ 3 is used in the experiment. The comparison results are presented in Table \ref{tbl_fpr}. In terms of the running time and FLOPs, UWCNN, NU2Net, TACL, and WaterNet rank the top four. The running time of all these four methods are all less than 0.04s. This indicates that their processing speed surpasses 25 frames per second (FPS), which is suitable for real-time underwater video processing. Our proposed CCL-Net holds a moderate position among all UIE methods. Even though our proposed CCL-Net consists of two stages, it is still faster than two single-stage deep UIE methods, i.e., UColor and PUIE, and all traditional methods. According to our experimental observation, CC-Net accounts for 90\% of the running time. Most of the computational time is currently dedicated to the repetitive operations involved in the FAB module. In the future, we plan to replace the FAB module with other high-efficiency attention modules and reduce the number of duplicate operations. In terms of the model parameters, our proposed CCL-Net ranks the second place, which is only larger than UWCNN. This indicates that CCL-Net is highly suitable for the embedded UAVs with limited storage space. 

\begin{figure}[!t]
\centerline{\includegraphics[width=3.2in]{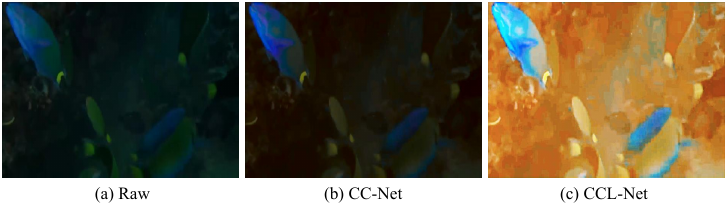}}
\caption{Unnatural colors introduced by CCL-Net on extreme low light underwater image.}
\label{fig_discussion_uc}
\end{figure}

\begin{figure}[!t]
\centerline{\includegraphics[width=3.2in]{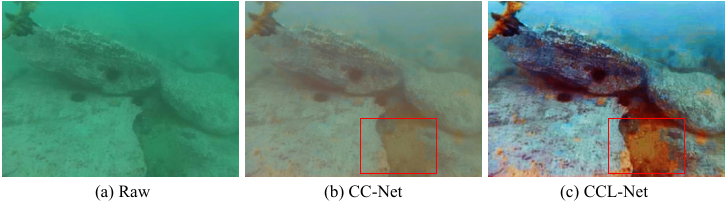}}
\caption{Inconsistent colors generated by CCL-Net on the same category or same object.}
\label{fig_discussion_ic}
\end{figure}

\subsubsection{Failure Cases}
Although our proposed CCL-Net shows pleasing visual quality and competitive quantitative performance, there are still several limitations. First, our method will introduce unnatural color artifacts when dealing with underwater images with extremely low illumination, as shown in Fig. \ref{fig_discussion_uc}. The potential reason is that the number of underwater images with extremely low illumination is relatively scarce in the training dataset or the quality of the corresponding pseudo-reference image is not sufficiently good. For future work, we plan to employ semi-supervised learning method which uses limited amount of high-quality underwater image pairs and a large number of underwater raw images without pseudo-reference images for training to improve the performance. Second, the colors of the same category or the same object are inconsistent, as shown in Fig. \ref{fig_discussion_ic}. We observe that this artifact is introduced by the limited performance of CC-Net, as shown in Fig. \ref{fig_discussion_ic} (b). In the future, we plan to introduce a pre-trained model which can generate underwater semantic features to guide the training of CC-Net.

\section{Conclusion}
This paper has presented a novel UIE method based on a two-stage framework with the assistance of cascaded contrastive learning. 
The key innovations are twofold. First, we propose a two-stage framework tailored to address the color cast and hazy effect issues in a divide-and-conquer manner. In this way, the complex underwater degradations can be better addressed with tailored network structures and loss constraints. Second, we apply contrastive loss as additional constraint to guide the training of each stage so as to guarantee the underwater image can be progressively enhanced. Comprehensive visual and quantitative comparative experiments on diverse benchmark datasets with various scenes and degradation types demonstrate the superiority, stability, and generalization ability of our proposed CCL-Net over the state-of-the-art comparative methods. Additional ablation experiments further verify the effectiveness of the core components in CCL-Net.

\bibliographystyle{IEEEtran}
\bibliography{Mybib}

\end{document}